\documentclass{article}

\usepackage{arxiv}

\usepackage[utf8]{inputenc} % allow utf-8 input
\usepackage[T1]{fontenc}    % use 8-bit T1 fonts
\usepackage{hyperref}       % hyperlinks
\usepackage{url}            % simple URL typesetting
\usepackage{booktabs}       % professional-quality tables
\usepackage{amsfonts}       % blackboard math symbols
\usepackage{nicefrac}       % compact symbols for 1/2, etc.
\usepackage{microtype}      % microtypography
\usepackage{lipsum}

\usepackage{lineno}
\usepackage{times}%Usa a fonte Latin Modern
\usepackage{indentfirst}%Indenta o primeiro parágrafo de cada seção.
\usepackage{nomencl}%Lista de simbolos
\usepackage{color}%Controle das cores
\usepackage{mathptmx}
\usepackage{siunitx}
\usepackage{longtable}
\usepackage{multirow}
\usepackage{makecell}
\usepackage{gensymb}
\usepackage{setspace}
\usepackage[ruled, linesnumbered]{algorithm2e}
\usepackage{cite}
\usepackage{amsmath,amssymb,amsfonts}
\usepackage{algorithmic}
\usepackage{graphicx}
\usepackage{textcomp}
\usepackage{xcolor}

\title{A Temporal Clustering Algorithm for Achieving thetrade-off between the User Experience and the Equipment Economy in the Context of IoT}

\author{
  Caio Ponte\\
  Programa de Pós Graduação em Informática Aplicada\\
  Universidade de Fortaleza\\
  Fortaleza, Brazil \\
  \texttt{caioponte@edu.unifor.br} \\
  \And
  Carlos Caminha\\
  Centro de Ciências Tecnológicas\\
  Universidade de Fortaleza\\
  Fortaleza, Brazil \\
  \texttt{caminha@unifor.br} \\
  \And
  Rafael Bomfim\\
  Programa de Pós Graduação em Informática Aplicada\\
  Universidade de Fortaleza\\
  Fortaleza, Brazil \\
  \texttt{rafaellpontes@gmail.com} \\
  \And
  Ronaldo Moreira\\
  Programa de Pós Graduação em Informática Aplicada\\
  Universidade de Fortaleza\\
  Fortaleza, Brazil \\
  \texttt{ronaldo@edu.unifor.br} \\
  \And
  Vasco Furtado\\
  Programa de Pós Graduação em Informática Aplicada\\
  Universidade de Fortaleza\\
  Fortaleza, Brazil \\
  \texttt{vasco@unifor.br} \\
}

\begin{document}
\maketitle

\begin{abstract}

We present here the \textit{Temporal Clustering Algorithm} \emph{(TCA)}, an
incremental learning algorithm applicable to problems of anticipatory computing
in the context of the Internet of Things. This algorithm was tested in a
specific prediction scenario of consumption of an electric water dispenser
typically used in tropical countries, in which the ambient temperature is around
30-degree Celsius. In this context, the user typically wants to drinking iced
water therefore uses the cooler function of the dispenser. Real and synthetic
water consumption data was used to test a forecasting capacity on how much energy
can be saved by predicting the pattern of use of the equipment. In addition to
using a small constant amount of memory, which allows the algorithm to be
implemented at the lowest cost, while using microcontrollers with a small amount
of memory (less than 1Kbyte) available on the market. The algorithm can also be
configured according to user preference, prioritizing comfort, keeping the water
at the desired temperature longer, or prioritizing energy savings. The main
result is that the \emph{TCA} achieved energy savings of up to 40\% compared to the
conventional mode of operation of the dispenser with an average success rate
higher than 90\% in its times of use.

\end{abstract}

% keywords can be removed

\section{Introduction}

The emergence of the Internet of Things ($IoT$) has enabled what is called
anticipatory computing. “Things”, while being active objects or even agents, are
increasingly possessing autonomy, which gives them the capability of
anticipating people’s wishes and intentions, as well as controlling their own
behavior, for example, optimizing their energy expenditure. This context of
$IoT$ finds users of technology that are increasingly demanding and that require
the intelligence of machines to provide customized answers to their demands
\cite{lytras2017}.

Although there is plenty of literature on $IoT$ applications that take into
account user preference \cite{guo2013,atzori2010}, there are few alternatives
that examine the issue from a conflicting multi-objective perspective. It is
common to come across problems in which the optimization of the behavior of the
“Thing” (here named object) can lead to an unsatisfactory user experience.
Providing tools that work from this perspective of multi-objectives is
fundamental, because they allow for a maximized trade-off between user
experience and efficiency of the behavior of the object.

Combined with this challenge of handling conflicting objectives, there is the
challenge to develop a solution that requires little memory (thus making its
installation and microcontrollers used inexpensive) and low processing power.
Specifically, the clustering of time series, in the context of
$IoT$, is a challenging problem because, at first, time series data is usually
much larger than the size of the memory and consequently is stored on disks
\cite{aghabozorgi2015time}. This leads
to an exponential decrease in the speed of the clustering process. Although the
evolution of microcomputers and servers is constant, where they are capable of
increasing processing and storage, the nature of the $IoT$ demonstrates the need
for a computational solution that is capable of performing in microcontrollers,
and/or in small computers, which often have severe processing and memory
restrictions \cite{zhang2017}. In addition, by investing in a low-cost technology
in household equipment, for example, it is possible to add the
market value in return for a small increase in manufacturing cost, allowing for
intelligent behavior to be added that will enhance the user experience
\cite{tan2018designing}.

This research aimed to evolve the behavior of a piece of domestic equipment,
specifically an electric water cooler and dispenser. This type of equipment is
often used in tropical countries, where the ambient temperature often exceeds
30\degree C (86\degree F). This type of dispenser, when in operation, activates
its electric compressor and cools the water inside a thermal tank, keeping it in
a temperature range, usually between 9 and 12\degree C. The compressor maintains
the temperature even at times when there is no water consumption, which can lead
to wasted energy, as there is energy expended to keep the water cold during
periods where it probably will not be consumed. Summarizing, although cooling is
essential for the water to be consumed at a pleasant temperature, the intense
use of the compressor of the dispenser generates greater power consumption. This
compromise between the experience of drinking cold water and the increase in the
energy bill is the focus of the discussion in this article. At one extreme (as
it is commonly known, since it is the standard on the market) one has to make
the compressor stay on continuously and so, every time the user goes to the
water to drink he/she will always receive properly cooled water. At the other
extreme, if the compressor only turns on a few times a day, energy savings will
occur, but the user's satisfaction in drinking chilled water will probably not
always occur.

In order to meet the above requirements, the main contribution of this work is
the development of an algorithm of incremental learning that seeks to discover
the pattern of use of the appliance so that it is possible to control the
electric water dispenser while maintaining the quality of the user experience
(here represented by the satisfaction of drinking chilled water). We propose
herein the \emph{Temporal Clustering Algorithm ($TCA$)}, an algorithm that
factors multiple time series in a set of non-overlapping segments, known as time
clusters. The agglomerative behavior of the $TCA$ is inspired by the \emph{City
Clustering Algorithm ($CCA$)} \cite{makse1998modeling,rozenfeld2008laws} a spatial
agglomeration algorithm widely used in defining cities beyond their boundaries
\cite{caminha2017human, caminha2017impact,rozenfeld2011area,
duranton2014growth}. The main feature of the $TCA$ is to consider user
preference through its three modes of \emph{Comfort}, \emph{Balance}, and
\emph{Eco}. \emph{Comfort} mode means that the algorithm will give a greater
weight to user satisfaction compared to energy savings. In \emph{Eco} mode the
opposite will occur, while \emph{Balance} mode will seek to balance the
importance of these criteria.

From tests with real and synthetic data of consumption of an electric water
dispenser, it was verified that the identification of these agglomerates is
useful to predict consumption schedules, allowing for the compressor to be used
in an intelligent way, favoring, according to the user’s preference, the economy
of energy or ensuring that chilled water is consumed even at times when the
usage of the appliance is infrequent. The energy saving using the $TCA$ was
compared with the conventional mode of the water cooler and dispenser. $TCA$ has
shown that it achieves energy savings of up to 40\% compared to this mode with
an average accuracy of more than 90\% of dispenser usage times. Its success rate
and memory consumption was also compared with some of the most commonly used
clustering algorithms in the literature and it was shown that the $TCA$ exceeds,
for the specific application scenario discussed in this article, these
algorithms in many situations.

\section{State of the Art}

Recent works also explore the combination of hardware and software for
monitoring and control in urban scenarios ideal for the use of the $IoT$
paradigm, such as the one addressed here \cite{hsu2018condition,tan2017wind}.
Orsi {\it et al} \cite{orsi2017smart} presented a system that integrated a
hardware controller for energy efficiency, a communication protocol to improve
data transmission, and a software module for planning and managing household
devices, which operates according to user preferences and maximum power
consumption. In Orsi's article, integration with machine learning for pattern
detection is a future work, which $TCA$ proposes to do.

Madiraju {\it et al} \cite{madiraju2018deep} proposed a new robust deep temporal
clustering algorithm based on Deep Temporal Clustering ($DTC$) to naturally
integrate dimensionality reduction and temporal clustering into a single
learning structure that is totally unattended. They claim that the clustering
layer of their algorithm can be adjusted to any temporal similarity metric, and
compares several similarity metrics and latest generation algorithms. The
viability of the algorithm is demonstrated using time-series data from several
domains, ranging from earthquakes to spacecraft sensor data. Despite the
importance of this study, due to the characteristic memory usage of Deep
Learning solutions in the training process, its application in the specific
context of this work is not feasible.

Finally, it is important to mention the work of Aghabozorgi {\it et al}
\cite{aghabozorgi2015time}. The authors make a rather complete survey proposing
a categorization of the main components that characterize the task of grouping a
time series. Despite the completeness of this and other studies
\cite{zhou2013hierarchical, quero2017dynamic}, the focus of the authors often
was on the efficiency and complexity of the approaches in the context of big
data and cloud computing. Our context is different because we seek to solve a
time series prediction problem with severe processing and storage limitations.

\section{Temporal Clustering Algorithm}

The $TCA$ receives as input a list of time series and composes an event density
prototype, $D$, which represents the average behavior of these series. This
prototype can be represented computationally from a vector of $N$ positions,
where an integer value is assigned to each position, $i$, of the vector, by
measuring the amount of events occurring at a given moment (determined by the
position $i$ of the vector) in $D$. The density of elements $D_i$, at each time
period $i$, is given by the mean of events occurring at $i$ in all the time
series.

The agglomerative behavior of the $TCA$ is inspired by the \emph{City
Clustering Algorithm (CCA)} \cite{makse1998modeling,rozenfeld2008laws}, a
spatial agglomeration algorithm widely used in defining cities beyond their
boundaries. The $CCA$ aggregates spatial units by considering two parameters,
one of which is a distance threshold and the other a threshold of population density.
The relationship between the $CCA$ and the $TCA$ is precisely in the use of these thresholds,
however, the $TCA$ uses them in only one dimension (temporal dimension).

\begin{figure}[!ht]
\centerline{\includegraphics[width=1.0\textwidth]{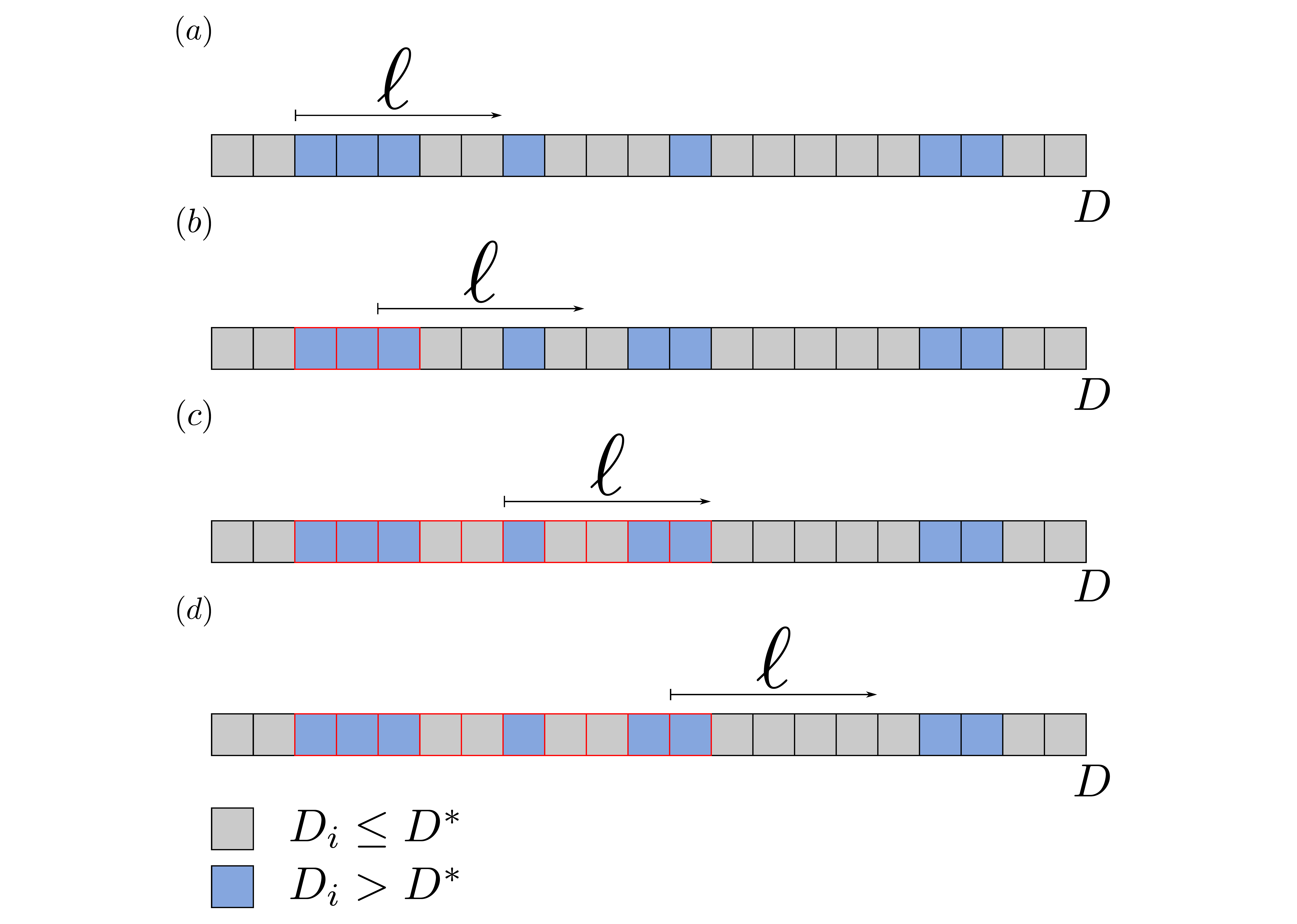}}
\caption{{\bf Cluster example found by the $TCA$ algorithm.}(a) The algorithm
starts from the first index of the vector that has a density of events, $D_i$,
greater than $D^*$ (populated index), then a time-length line $\ell$ is plotted
which verifies the subsequent indexes with $D_i>D^*$ that occurred within this
time limit. (b) From the last populated index of the vector found within the
established threshold, a line of length $\ell$ is drawn again. (c) The process
is repeated until, in (d), no populated index is found with a distance smaller
than $\ell$, at that time a cluster is found and the algorithm returns to the
initial step to find the next cluster.}
\label{esquema-tca}
\end{figure}

The $TCA$ makes use of two thresholds. The first, $\ell$, a time threshold,
represents the distance between the elements of the time series to consider them
as temporally contiguous, more precisely, all events that are at temporal
distances smaller than $\ell$ are grouped together. The second, $D^*$, is an
event density threshold, used to consider the agglomeration of low density
elements in the series, defined according to the user’s preference. The density
of events $D_i$ is located at each index $i$ of the vector, if $Di>D^*$, then
the index $i$ is considered populated, and consequently, it can be grouped. From
the definition of the values of $\ell$ and $D^*$, the behavior of the $TCA$ in
the process of identifying a cluster can be observed schematically in Figure
\ref{esquema-tca}.

The $TCA$ still uses a percolation model \cite{stauffer1994} to choose the
threshold value $\ell$. The size of the largest cluster is measured while the
threshold $\ell$ is in search of a phase transition. More precisely, it ensures
the moment when the largest cluster aggregates the second largest cluster, which
is said to be the critical point of the system. The threshold value should be
chosen at the moment before the phase transition, at which time the clusters are
consolidated and the minimum increase in the value of $\ell$ causes them to come
together.

By evaluating the density of water withdrawals from the electric water cooler
and dispenser, it is possible to define three modes of operation for the $TCA$:
\emph{Comfort}, \emph{Balance}, and \emph{Eco}. The \emph{Comfort} mode
prioritizes the quality of the experience, while the \emph{Eco} mode gives
priority to the energy saved, finally, the \emph{Balance} mode aims at a balance
between comfort and economy. The values of $D^*$ can be associated with the
proposed modes of operation from the observation that the time series
distribution used in the evaluation is exponential ($\mu \approx \sigma$). From the analysis of real and synthetic data on water consumption patterns (more details in \ref{discussion}), we propose: $D^*=\mu$ for \emph{Eco} mode, $D^*=\mu-\sigma/2$ for \emph{Balance} mode and
$D^*=0$ for \emph{Comfort} mode.

Algorithm \ref{alg:alg1} illustrates steps of the $TCA$ using as input a
list of temporary series, the period in minutes (which corresponds to the time
of each $D_i$) and the operation mode, defined by the user as a function of their
choice whether to prioritize the lowest power consumption or to drink cold water
more frequently. The prototype, $D$, is constructed at lines 2 and 3. From lines
4 to 9 it is checked which $D^*$ value should be chosen as a function of the
mode used as a parameter. The percolation process is performed by the loop that
starts at line 10 and ends at line 17. The detection of clusters from a fixed
value of $\ell$ and $D^*$ is performed between lines 11 and 15. It is noteworthy
that $TCA$ has a dynamic behavior, essentially when applied to problems of
forecasting times of use of home appliances, where the user can use the
parameter $D^*$ to define configurations that prioritize the energy saving to
the detriment of the accuracy in the forecast of use. We will show that from the
verification that the usage data of the appliance follow a certain distribution,
it is possible to define modes of operation for the equipment. Specifically, we
show how the $D^*$ values should be associated to three modes of operation:
\emph{Comfort}, \emph{Balance}, and \emph{Eco}, where the last prioritizes the
energy saving, the first the comfort (higher rates of accuracy in forecasts) and
the second a balance between comfort and energy saving. In Section
\ref{discussion} we will discuss the impact of the $D^*$ values on the energy
savings provided by the $TCA$.

Due to the need to store the vector elements saving the time series data with
$D_i>D^*$ and the clusters found, the spatial complexity of the $TCA$ for the
fixed threshold value $\ell$ is $O(N+k)$, where $N$ is the number of elements in
the time series and $k$ is the number of clusters found. With respect to
temporal complexity, we have a complexity $O(N)$, since it will be necessary to
go through all the elements of the prototype of density of events, $D$, only
once to find the clusters.

Considering the percolation model, the spatial complexity of the $TCA$ remains as
$O(N+k)$, since with each increase in $\ell$ it is not necessary to store the
agglomerates found. Only two variables will be added at spatial cost, which will
store the value of the largest jump in the size of the largest cluster and the
moment of the phase transition. With respect to time complexity, we have an
increase for $O($ $N$ $x$ $\omega)$, where $\omega$ refers to the amount of
increases that will be made in $\ell$ during the execution of the percolation
model.

\begin{algorithm}
   \SetAlgoLined
   \KwData{$series, period, mode$}
   \Begin{
   
   $D = \{ \emptyset \}$ \\
   $\{ D_i \}$ = mean of the values of each index, $i$, in $series$ \\

\uIf{$mode$ == ``Comfort''}{
      $D^*=0$
   }
   \uElseIf{$mode$ == ``Balance''}{
      $D^*=\mu-\sigma/2$, where $\mu$ and $\sigma$ are the mean and standard deviation of $D$, respectively \\
   }
   \uElseIf{$mode$ == ``Eco''}{
      $D^*=\mu$, where $\mu$ is the mean of the values of $D$ \\
   }
	   
\For{$\ell \leftarrow period$ \KwTo $size(series)$ \textbf{\textsc{by}} $period$}{       	     	
       	\While{ there exist elements in $D$ with $D_i>D^*$}{	  
	  		\uIf{ $\ell$ reaches some event with $D_i>D^*$}{	
			shift $\ell$ to the period in which the last clustered event occurs \\
	  		}	  
	  		create a new cluster	  
       	}
   }   
   $clusters \leftarrow $ get the clusters to $\ell$ at the moment prior to the critical point   
   \Return clusters   
   }   
   \caption{\textsc{Temporal Clustering Algorithm}}\label{alg:alg1}
\end{algorithm}

\section{Benchmark}

Three datasets were used to perform accuracy tests of TCA and three other
reference algorithms in terms of their respective energy consumptions, in $Wh$
per day. The datasets inform the times of the day when there was withdrawal of
water. The first dataset was obtained by monitoring the water consumption from a
test electric water dispenser in a corporate environment. In the environment in
question, 16 employees consumed water from the appliance over 5 business days,
specifically from 06 to 12 December (days 09 and 10 were excluded because they
were not working days). In all, 301 water withdrawals were carried out from the
appliance, with an average of 60.2 withdrawals per day. It is noteworthy that,
because it was test equipment, there was an agreement between the factory that
produced the electric water dispenser (with the inserted $TCA$) and the company
where the appliance could be tested for only 5 days. For this reason, only
five-day real data was used in our test.

The remaining datasets, a commercial synthetic data ($CS$) and another, a
residential synthetic data ($RS$), were generated following an exponential
distribution \cite{peebles2001probability} to simulate the rate of use of the
electric water dispenser in a corporate and domestic environment, respectively.
Synthetic data was generated through a value-generating function that follows
an exponential distribution and such values represent the time difference
between each consecutive event. The function of the probability density of an
exponential depends just on the average of the random variable, which in this
case is the time difference between drinks. In choosing the average to be used
in the data generating function, we took into account the division of the day
into shifts and the application of different averages for each shift according
to the pattern of use expected during the day. For example, in a corporate
environment the most intensive use of withdrawals is expected during the
commercial period with longer breaks between the lunch period while in a
residential setting the most constant use is expected throughout the day but
with less frequency.

For the application of the $TCA$, the data sets were treated so as to accumulate
the amount of water withdrawals ($D_i$) in ten-minute intervals ($period = 10$)
in Algorithm \ref{alg:alg1}, that is, for each day there are 144 observations,
where each observation is associated with the number of withdrawals of water.
The other algorithms used in the comparisons perform their processing with the
original data of the time of drinks. 
%Figure \ref{histograma} shows the histogram
%of the $D_i$ value for the five days of the real dataset. The dashed curve
%represents the logarithmic regression that best fits the data and has a standard
%coefficient error of 0.037.

In the specific case of clustering algorithms, what will be interpreted is when
the found agglomerations determine the moments when the water of the appliance
should be ice cold. In this context, the accuracy of the algorithm will be
increased whenever an event of a test time series occurs within one of the
identified clusters, in a complementary way, whenever an event occurs outside
the limits of one of the clusters, the error rate of the algorithm will be
increased. In other words, the error rate is the percentage of water requests
that were made in periods that were outside some cluster found by the algorithm.
At the same time, the energy consumption will be measured by the time in which
it will be necessary to maintain the compressor of the connected electric water
dispenser. In all the tests the compressor will be connected within the
estimated time clusters and will be adding $1.62kW$ per hour while the
compressor remains turned on. In order to help the monitoring of the functioning
of $TCA$, we have developed a simulator in $Python$, version 3.6, which allows
to visualize instantly the energy consumption and the error rate. The simulator
helps also to visualize the clusters found by TCA. The details of the simulator
are described in the supplementary material.

In our tests, in addition to the $TCA$, the standard electric water dispenser
performance algorithm, herein called \emph{Conventional}, was used, and the
\emph{K-Means} and \emph{Expectation- Maximization (EM)} algorithms were used.
The \emph{Conventional} algorithm does not save energy, it keeps the water in the appliance
chilled all the time. \emph{K-Means} is a clustering algorithm that agglomerates
observations (in the case of this work each observation is a time of day in
which a drink was taken), receiving as a parameter the quantity of clusters $k$
that one wants to find \cite{hartigan1979algorithm}. Considering that there are
four shifts in one day, $k=4$ was used in the tests of this algorithm. Finally,
$EM$ is an iterative algorithm to find maximum likelihood estimates of
parameters in statistical models, where the model depends on unobserved latent
variables \cite{moon1996expectation}. It should be noted that the
$TCA$ is an innovative algorithm that proposes to solve a very specific problem,
considering the user's preference, so it is important to find algorithms that
can be compared with it in all its functions. In this section, our main objective
is to show that the proposed algorithm has a similar accuracy to some classic
clustering algorithms. When comparing with the \emph{Conventional} mode, our goal is
to show how much energy can be saved by adding some intelligence to the
compressor behavior of the electric water dispenser.

Due to the memory limitation imposed by the application context of the
algorithm, in all scenarios, to conduct the accuracy tests, two days of training
data and another day of test data were used, separating the days into intervals
of ten minutes, that is, $period=10$ was used as input in the $TCA$ clustering
module (test varying $period$ and the number of time series put into the $TCA$
are present in the supplementary material). For the other algorithms, the
standard of two of training and one of test are also used, however, they perform
the clustering process using the original data of the water withdrawal times,
therefore, each sample represents the water withdrawal moments. In all tests
a cross validation was performed, using real and synthetic data, combining any
two days of training with another test day.

For the $TCA$, in all the tests performed in this section, we assume $D^*=0$. In
this configuration the algorithm assumes a mode of operation where comfort is
prioritized, which in the context of the applied scenario of this work means
that the possibility of the user to consume unrefrigerated water will be
minimized.

\begin{table}[htbp]
\centering
\caption{Memory consumption (bytes), average energy consumption (Wh) per day and average algorithm error}
\label{my-label}
\begin{tabular}{cccccccc}
\Xhline{2\arrayrulewidth}
\multirow{2}{*}{\textbf{}} & \multirow{2}{*}{\textbf{\begin{tabular}[c]{@{}c@{}}Memory\\ (Bytes)\end{tabular}}} & \multicolumn{3}{c}{\textbf{Consumption (Wh)}} & \multicolumn{3}{c}{\textbf{Error}}        \\
                           &                                                                                    & \textbf{Real}   & \textbf{CS}   & \textbf{RS}  & \textbf{Real} & \textbf{CS} & \textbf{RS} \\ \Xhline{2\arrayrulewidth}
\textbf{TCA}               & $\approx 300$                                                                      & 10.98          & 13.12        & 13.19       & 0.14        & 0.03      & 0.10      \\
\textbf{K-Means}           & $\approx 2900$                                                                     & 10.73          & 13.04        & 13.02       & 0.27        & 0.08      & 0.14      \\
\textbf{EM}                & $\approx 2900$                                                                     & 10.55          & 13.34        & 12.78       & 0.31        & 0.04      & 0.06      \\
\textbf{Conventional}      & -                                                                                  & 14.64          & 14.75        & 13.25       & 0.00        & 0.00      & 0.00      \\ \Xhline{2\arrayrulewidth}
\end{tabular}
\label{table1}
\end{table}

Table \ref{table1} shows the result of the comparison. In all the studied
algorithms it is possible to observe the memory consumption, energy, and the
error rate. The memory consumption was estimated according to
the input of each of the algorithms. In the case of the $TCA$ this input is fixed,
there are two vectors of 144 bytes that added to the control variables up to
300 bytes. In regard to $EM$ and \emph{K-Means}, in the worst case, they can receive
infinite entries in one day. A more realistic view would be to assume the
existence of an event per minute as the worst case. Thus, by adding two
vectors of 1440 positions, plus control variables, the use of memory
at $\approx 2900$ bytes was considered. In relation to energy consumption, even in its least
economical mode, the $TCA$ had similar consumption to \emph{K-Means} ($k=4$) and
$EM$ algorithms, however, the $TCA$ had less error in almost all the tests.
Although $EM$ has consumed less energy, the $TCA$ may sacrifice some of its
accuracy to decrease consumption. Section \ref{discussion} looks at the
consequences of changing the value of $D^*$.

\section{TCA and the User Preference}\label{discussion}

The $TCA$ obtained promising success rates using the least amount of memory
among all the algorithms tested. It is worth noting that, considering the
restrictions imposed by the application scenario, it was the only algorithm
capable of using less than 512 bytes of memory, allowing it to be used by some
of the cheapest microcontrollers on the market.

\begin{figure}[!ht]
\centerline{\includegraphics[width=1.0\textwidth]{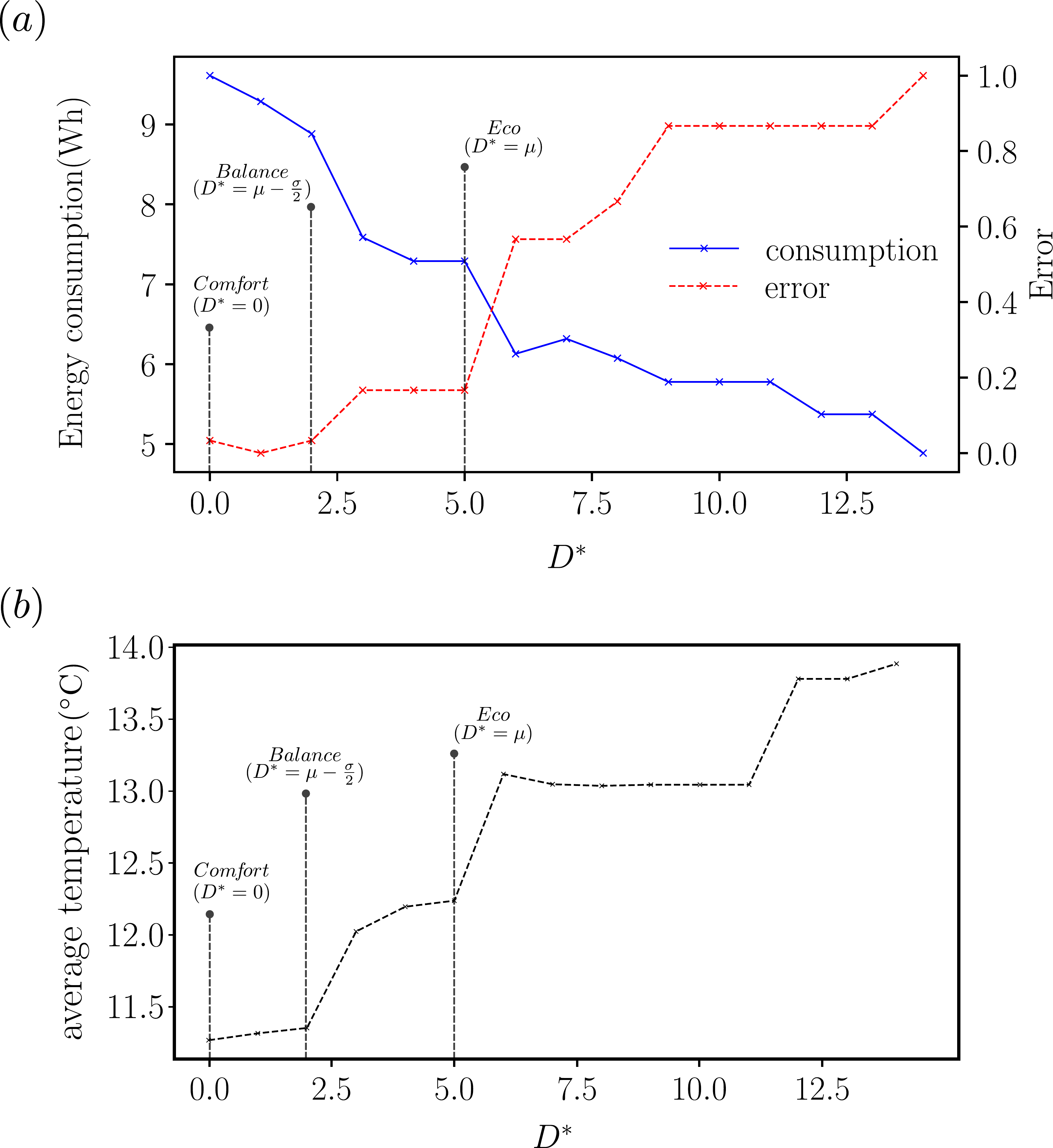}}
\caption{{\bf Impact of choosing the $D^*$ parameter on the quality of the user
experience in an experiment with real data.} (a) The solid blue line illustrates
the behavior of energy consumption and the red dotted line shows the error rate
as the value of $D^*$ is increased. (b) The black dotted line shows the mean
temperature of the water at the time of consumption as a function of the choice
of threshold $D^*$. In both figures the values of $D^*$ are demarcated in the
three proposed modes of operation.}
\label{aval-d}
\end{figure}

Although these results alone are presented as encouraging, the great
differential of the $TCA$ lies in its dynamic behavior, where, according to the
user’s preference, the algorithm can prioritize energy savings, by not
aggregating time slots where consumption is below the $D^*$ threshold, or
maximize the quality of user experience, by considering the aggregation of time
slots where water consumption is very low. In other words, by manipulating the
parameter $D^*$ the user can inform the algorithm of what s/he considers as a
low density of temporal events (\emph{e.g.}, water withdrawals from a water
dispenser), and consequently define certain moments of the day where, due to the
low frequency of use of the appliance, it is acceptable to drink water at a
higher temperature in order to save energy. Figure \ref{aval-d}(a) illustrates
the error rate behavior and energy consumption by varying the $D^*$ parameter in
an experiment with the real data. It is observed that, as the value of $D^*$
increases, the energy consumption decreases and the error rate increases.

It is also possible to observe the relationship between the parameter $D^*$ and
the quality of the user’s experience, measured indirectly by the average
temperature of the water consumed. In Figure \ref{aval-d}(b) it is observed that
the increase of $D^*$ occurs at the cost of consumption at a higher average
temperature, revealing a relation between $D^*$ and the quality of the
experience.

Therefore, the proposal is to use $D^*=\mu$ for \emph{Eco} mode,
$D^*=\mu-\sigma/2$ for \emph{Balance} mode and $D^*=0$ for \emph{Comfort} mode.
In our tests we observed that for $D^*=\mu-\sigma/2$ the error rates remain
low, always below 8\% and decreasing by approximately 1 $Wh$ in relation to
$D^*=0$ (in the scenario illustrated in Figure \ref{aval-d}(a) 7.2\% of error
rate was obtained with 8.9 $Wh$ of consumption in \emph{Balance} mode). For the
\emph{Eco} mode ($D^*=\mu$), the error rates remained below 20\%, with up to 2.5
$Wh$ less consumption than \emph{Balance} mode ($D^*=\mu-\sigma/2$).

Table \ref{tbl-modos-operacao} shows the mean error rate and mean energy
consumption in tests with real and synthetic data. It is observed that the
change in operating mode has a direct impact on energy savings, measured in
$Wh$. Similarly, it is observed that the lower the average consumption, the
higher the average error rate.

\begin{table}[htbp]
\centering
\caption{Average energy consumption (Wh) per day and mean error for the modes of operation of the $TCA$.}
\label{tbl-modos-operacao}
\begin{tabular}{ccccccc}
\Xhline{2\arrayrulewidth}
\multirow{2}{*}{} & \multicolumn{3}{c}{Consumption (Wh)} & \multicolumn{3}{c}{Error} \\
                  & Real        & CS         & RS         & Real    & CS     & RS     \\ \Xhline{2\arrayrulewidth}
\textbf{Eco}               & 8.36       & 11.39     & 11.41     & 0.34  & 0.16 & 0.29 \\
\textbf{Balance}           & 10.03      & 12.07     & 13.16     & 0.13  & 0.06 & 0.10 \\
\textbf{Comfort}           & 10.98      & 13.12     & 13.19     & 0.14  & 0.03 & 0.10 \\ \Xhline{2\arrayrulewidth}
\end{tabular}
\end{table}

In order to evaluate the effectiveness of our approach, we have conducted new
tests (pilot study), during one month, with a dispenser prototype in a corporative
environment. All the tests were made with the prototype in the mode {\it Comfort}. The
results from the tests with the prototype have shown that the average
consumption of the prototype using $TCA$ is $9.81$ $Wh$. This means a 32\% of
economy compared with an equipment without $TCA$. During this period in 90\% of
the times that the user accessed the water cooler, the water was cooled in the
right temperature.

\section{Conclusion}

This paper presents a new algorithm for Temporal Clustering, known as the
\emph{Temporal Clustering Algorithm}. The behavior of the algorithm in question
is inspired by the \emph{City Clustering Algorithm}
\cite{makse1998modeling,rozenfeld2008laws}, an algorithm widely used in problems
with definitions of city boundaries. The great differential of the $TCA$ is in
its ability to combine significant rates and prediction with a dynamic behavior,
where it is possible to be configured according to the user’s preference. It was
also shown that the algorithm uses little memory, making it possible to
incorporate it into low-cost circuits, which are constantly used in applications
of the Internet of Things. Real and synthetic data of water consumption of an electric water dispenser was used to evaluate the level of prediction of the $TCA$. In particular, the
performance of the algorithm was verified in function of its ability to predict
periods where the appliance is being used, consequently verifying how much
energy the algorithm can save by turning off the electric water dispenser
compressor at times when it is not in use. A comparison with some of the main
algorithms in the literature has shown that the $TCA$ can save energy (even in
its least economical configuration), use less memory and obtain greater
accuracy.

\end{document}